# DEVELOPMENT OF A 3D TONGUE MOTION VISUALIZATION PLATFORM BASED ON ULTRASOUND IMAGE SEQUENCES

*Kele Xu[1,2*], Yin Yang[3], A. Jaumard-Hakoun[1,2*], C. Leboullenger[1,2*], G. Dreyfus[2], P. Roussel[2*], M. Stone[4], B. Denby[1,2*]*

[1]Université Pierre et Marie Curie; [2]Signal Processing and Machine Learning Laboratory, ESPCI-ParisTech; [3]University of New Mexico; [4]University of Maryland Dental School; *Present affiliation: Langevin Institute, ESPCI-ParisTech
kelele.xu@Gmail.com, bruce.denby@Gmail.com

**ABSTRACT**

This article describes the development of a platform designed to visualize the 3D motion of the tongue using ultrasound image sequences. An overview of the system design is given and promising results are presented. Compared to the analysis of motion in 2D image sequences, such a system can provide additional visual information and a quantitative description of the tongue's 3D motion. The platform can be useful in a variety of fields, such as speech production, articulation training, etc.

## 1. INTRODUCTION

A silent speech interface (SSI) [1] is a system to enable speech communication with non-audible signals, that employs sensors to capture non-acoustic features for speech recognition and synthesis. Extracting robust articulatory features from such signals, however, remains a challenge. As the tongue is a major component of the vocal tract, and the most important articulator during speech production [2], a realistic simulation of tongue motion in 3D can provide a direct, effective visual representation of speech production. This representation could in turn be used to improve the performance of speech recognition of an SSI, or serve as a tool for speech production research and the study of articulation disorders.

Significant activity on modelling the dynamic 3D tongue, appears in the literature [3], [4], [5] and [6]. Currently, existing state-of-the-art platforms (such as ArtiSynth [6]) focus on modelling driven by muscle activations. Nevertheless, despite many attempts to characterize the bio-mechanical properties of the tongue [7], our understanding of tongue muscle activations is still incomplete. Furthermore, most existing 3D tongue visualization frameworks are unable to simulate real-time tongue motion. To address these challenges, a platform is being developed in our laboratory to visualize the tongue in real-time using ultrasound imaging, a relatively non-invasive and inexpensive modality. Although our primary goal is to design a platform for ultrasound data, the system can also serve as an interface for other imaging modalities (e.g., Magnetic Resonance Imaging (MRI)) to assist studies of speech production.

The article gives an overall description of the platform, which is still under development, and presents some promising initial results. The paper is organized as follows. In section 2, the ultrasound data acquisition and overview of the platform are outlined. Sample visualization results are given in section 3. A conclusion and some perspectives are presented in section 4.

## 2. OVERVIEW OF THE PLATFORM

The operation of the platform can be divided into four modules: data acquisition and pre-processing; subject-specific modelling; motion estimation (or tracking); and real-time dynamic modelling guided by the motion information.

### 2.1 Data acquisition and processing

A data acquisition helmet holds an ultrasound probe to capture the motion of the tongue at a frame rate of 60 Hz. The ultrasound machine chosen is the Terason T3000, which is lightweight and portable while retaining high image quality, and allows data to be exported to a portable PC via Firewire (further details appear in [8]).

### 2.2 Subject specific modelling

The generic tongue model used is that of ArtiSynth [6]; however, for a smoother mesh and better visualization results, the generic model was subdivided into roughly 13,000 nodes and 44,000 tetrahedral elements (as shown in Fig. 1). The platform provides a tool to adapt the tongue's rest shape in order to perform subject-specific modelling. A subject's tongue rest shape information can be obtained by combining midsagittal and coronal ultrasound scans. Surface nodes on the 3D tongue model are selected in the midsagittal plane and coronal plane. The subject-specific mesh editing is formulated as a contour-to-contour problem, and treated with an active contour model [9]. Nodes on the 3D tongue model's surface can be regarded as the control points of the spline, which are then "attracted" by the manually extracted coronal and midsagittal contours (as shown in Fig. 2.). The displacements of the nodes control the movements of the active contour model, which are used to guide the geometric change of the model tongue to fit the subject's tongue shape. The mesh editing processing can be described with linear constraint equations using the Lagrange multiplier method [10].

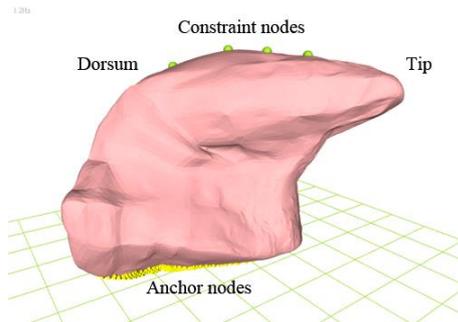

**Figure 1**: Generic tongue model with anchor (yellow) and mid-sagittal constraint nodes (green), for driving the model, are shown in the rest configuration. Anchor nodes' displacements are zero during the motion of the tongue model.

**Figure 2:** Subject-specific modeling. Here, the green points represent control points, which are attracted by the contours manually extracted (blue) from the coronal and midsagittal scan.

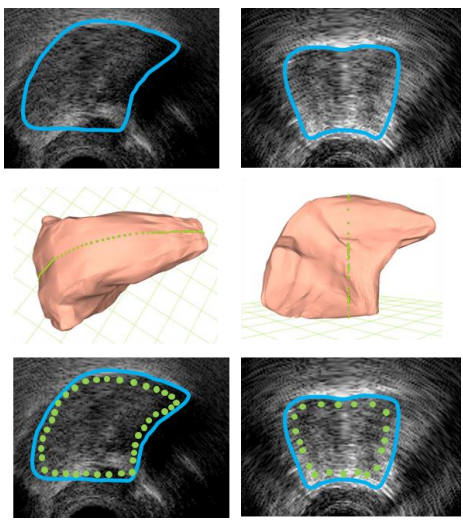

### 2.3 Motion estimation

Two approaches are used to track the motion of the tongue: speckle tracking and contour tracking.

In an ultrasound image, the acoustic scattering produces a pattern of gray values referred to as a speckle pattern [11], which, for a sufficiently high frame rate (the Terason here is clocked at 60 frames per second or fps), are preserved between subsequent image frames. Tracking characteristic speckle patterns can thus provide markers for tagging the soft tissue motion [12]. If the position of a segment of the tongue tissue changes, one may assume that the position of its acoustic fingerprint (texture) will change accordingly. By tracking these patterns, we can follow the motions of tissue in real-time. In our framework, the method of large displacement optical flow [13] is used to obtain correspondences between the patterns in 2D image sequences, from which the tongue's motion is derived. Although somewhat unstable, the technique provides a simple method for initial tests of the 3D visualization platform.

A alternative method to track motion of the tongue is to extract a tongue contour [14]. A new contour tracking technique has been developed and will be implemented in the platform in the near future [15]. The algorithm, rather than exploiting prior shape information obtained from other imaging modalities, extracts such information from adjacent images in the ultrasound sequence itself themselves, and uses it as an extra "force" to guide the tracking of the contour, so as to better handle missing or faint contours [16]. The algorithm has been shown to be quite stable on real ultrasound image sequences lasting up to 3 minutes.

### 2.4 Real-time dynamic modeling

The motion information extracted is subsequently transmitted to selected nodes on the midsagittal tongue model surface in order to drive the 3D model at the acquisition rate of the ultrasound image sequence. The displacements of the nodes obtained from the speckle tracking are applied as a linear constraint, which can be integrated into the governing equation of the dynamic deformable tongue model using the Lagrange multiplier method [10].

This can be expressed as:
$$\mathbf{M}\ddot{\mathbf{u}} + \mathbf{C}\dot{\mathbf{u}} + \mathbf{K}\mathbf{u} = \mathbf{f} \quad (1)$$

The mass, damping, and stiffness matrices $\mathbf{M}$, $\mathbf{C}$, $\mathbf{K}$ (of size $3n \times 3n$, where $n$ is the number of nodes) are determined by the material's intrinsic physical properties; $\mathbf{u}$ is the vector of the displacements of the nodes from their original positions on the mesh; **f** is the vector of external forces. Eq. (1) is a coupled system of ordinary differential equations, which typically cannot be solved in real-time. To address this problem, we adopted linear modal analysis to accelerate the computational efficiency by solving the generalized eigenproblem. Suppose $\mathbf{\Phi}$ and $\mathbf{\Lambda}$ (a diagonal matrix of eigenvalues) are the solution matrices for $\mathbf{K\Phi} = \mathbf{M\Phi\Lambda}$, where $\mathbf{\Phi^T M \Phi} = \mathbf{I}$, $\mathbf{\Phi^T K \Phi} = \mathbf{\Lambda}$. The modal displacement can be expressed as a linear combination of the columns of $\mathbf{\Phi}$.

$$\mathbf{u} = \mathbf{\Phi q} \qquad (2)$$

Substitution of Eq. (2) into Eq. (1) followed by a left-multiplication by $\mathbf{\Phi^T}$, results in:

$$\ddot{\mathbf{q}} + \mathbf{\Phi^T C \Phi} \dot{\mathbf{q}} + \mathbf{\Lambda q} = \mathbf{\Phi^T f} \qquad (3)$$

We can only use several dominant columns in $\mathbf{\Phi}$ (e.g. the ones associated with the smallest eigenvalues), thus the computation load of Eq. (3) is considerably reduced. Under the commonly adopted Rayleigh damping condition: $\mathbf{\Phi^T C \Phi} = \xi \mathbf{I} + \zeta \mathbf{\Lambda}$ (where $\xi$ and $\zeta$ are scalar weighting factors), the calculation can be carried out in nearly real time.

As the linear modal analysis cannot deal with large magnitude deformations, a modal warping technique [17] is used to compute the nonlinear deformation term, so that the new representation of the rotational part becomes:

$$\mathbf{w} = \mathbf{W\Phi q} \qquad (4)$$

where **w** is a vector representing the angular velocities of the nodes and **W** is the *curl* of the linear displacement. The Newmark average-acceleration method is used and Eq. (3) can be written as a simple linear system of:

$$\mathbf{A\ddot{q}} = \mathbf{b} \qquad (5)$$

which is to be solved repeatedly at each time step. More detailed formulation and derivation can be found in [18].

## 3. VISUALIZATION RESULT

Fig. 3 provides an overview of the user interface developed to implement the described platform. The 3D visualization strategy provides enriched awareness of tongue movement, and the potential benefits for clinical and research applications are clear.

Although only four constraint nodes are used to describe the motion of the tongue in the ultrasound image, the deformation simulated with the proposed framework is informative. Some results of the visualization platform on different vocalizations are shown in Fig. 4.

**Figure 3:** A snapshot of the user interface of the platform being developed

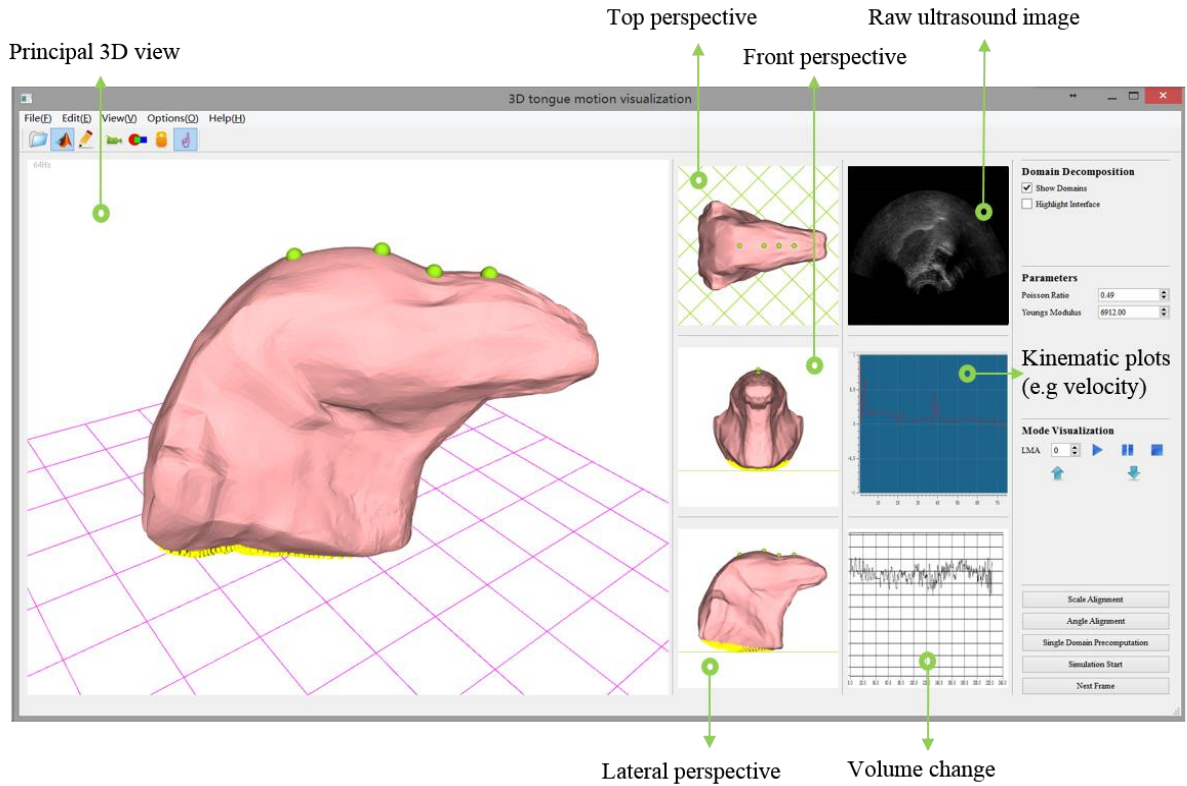

**Figure 4:** Some examples of the visualization results

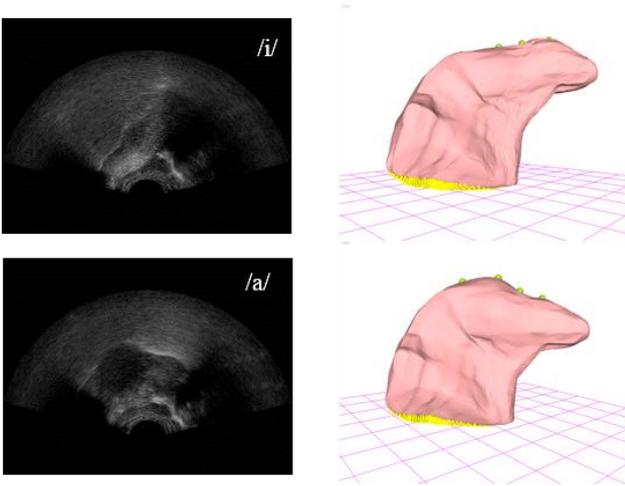

At present a volume-preservation constraint has not been included to the framework. Nevertheless, to test the change of tongue volume during deformations, an experiment was performed, with results as shown in Fig. 5. It can be seen that the change in the tongue volume remains small, less than 2% in most cases.

**Figure 5:** Volume change of the tongue model

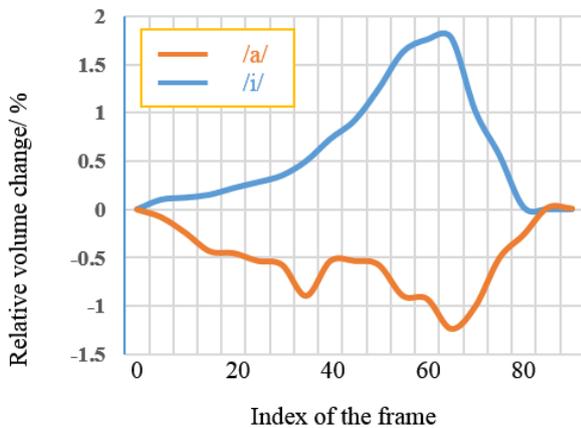

The described dynamic 3D tongue motion visualization platform is implemented in Microsoft Visual C++ 2010 in a Windows 7 environment, using a PC with Intel Core i7, 8G DDR3L and an NVIDIA GHTX862M. Using 150 modal bases (the number of the columns in matrix $\Phi$), the simulation throughput is 43.2 fps. Thus although tongue deformation during speech can be rapid, the developed framework appears to be able to meet this demand, following the motion and generating tongue shape in real-time with relatively good accuracy.

## 4. CONCLUSION

Speech is perhaps the most important human bio-signal, but not all of the characteristics of speech production are fully understood. As a new tool for understanding speech production, the proposed 3D tongue motion visualization platform has been developed, based on ultrasound images, using modal analysis to perform the simulation in real time. We believe this to be the first combination of ultrasound imaging with a 3D tongue model to visualize motion in real time. Our experimental results reveal promising functionality and potential applications in several fields.

There are still many improvements to be made. Due to the relatively low performance of point tracking with optical flow, manual refinements are required at present. Also, registration between 2D points in the ultrasound image and the nodes on the 3D tongue model is still under development, and manual tweaking is required here as well. Upcoming versions of the platform will include an improved real-time contour detection algorithm. An explicit volume-preserving constraint is also to be added.

## 5. ACKNOWLEDGEMENT

This work was partially funded by the European FP7 i-Treasures project (Intangible Treasures - Capturing the Intangible Cultural Heritage and Learning the Rare Know-How of Living Human Treasures FP7-ICT-2011-9-600676-i-Treasures). Kele Xu would also like to thank the China Scholarship Council.

## 6. REFERENCES

[1] Denby, B., Schultz, T., Honda, K., Hueber, T., Gilbert, J. M., Brumberg, J. S. 2010. Silent speech interfaces. *Speech Communication*, 52(4), 270-287.
[2] Stone, M. 2005. A guide to analysing tongue motion from ultrasound images. *Clinical linguistics & phonetics*, 19(6-7), 455-501.
[3] Stavness, I., Lloyd, J. E., Fels, S. 2012. Automatic prediction of tongue muscle activations using a finite element model. *Journal of Biomechanics*, 45(16), 2841-2848.
[4] Vogt, F., Lloyd, J. E., Buchaillard, S., Perrier, P., Chabanas, M., Payan, Y., Fels, S. 2006. Efficient 3D finite element modeling of a muscle-activated tongue. *Lecture Notes in Computer Science*, 4072, 19-28.
[5] Steiner, I., Ouni, S. 2011. Towards an articulatory tongue model using 3D EMA. *Proc. 9th ISSP* Montreal, 147-154.
[6] Lloyd, J. E., Stavness, I., Fels, S. 2012. ArtiSynth: A fast interactive biomechanical modeling toolkit combining multibody and finite element simulation. *Soft tissue biomechanical modeling for computer assisted surgery*. Springer, 355-394.

[7] Gérard, J. M., Ohayon, J., Luboz, V., Perrier, P., Payan, Y. 2004. Indentation for estimating the human tongue soft tissues constitutive law: application to a 3D biomechanical model. *Lecture Notes in Computer Science*, 3078, 77-83.

[8] Al Kork, S.K., Jaumard-Hakoun, A., Adda-Decker, M., Amelot, A., CrevierBuchman, L., Chawah, P., Dreyfus, G., Fux, T., Pillot, C., Roussel, P., Stone, M., Xu, K., and Denby, B. *Proc. 10th ISSP Cologne.* 5-8.

[9] Kass, M., Witkin, A., Terzopoulos, D. 1988. Snakes: Active contour models. *International Journal of Computer Vision*, 1(4), 321-331.

[10] Yang, Y., Rong, G., Torres, L., Guo, X. (2010). Real-time hybrid solid simulation: spectral unification of deformable and rigid materials. *Computer Animation and Virtual Worlds*, 21(3-4), 151-159.

[11] Meunier, J., Bertrand, M. 1995. Ultrasonic texture motion analysis: theory and simulation. *IEEE Transactions on Medical Imaging*, 14(2), 293-300.

[12] D'hooge, J. 2007. Principles and different techniques for speckle tracking. *Myocardial imaging tissue Doppler and speckle tracking*, Wiley, 17-25.

[13] Weinzaepfel, P., Revaud, J., Harchaoui, Z., Schmid, C. 2013. DeepFlow: Large displacement optical flow with deep matching. *Proc. of IEEE ICCV* Sydney, *1385-1392.*

[14] Li, M., Kambhamettu, C., Stone, M. (2005). Automatic contour tracking in ultrasound images. *Clinical linguistics & phonetics*, 19(6-7), 545-554.

[15] Xu, K., Yang, Y., Jaumard-Hakoun, A., Dreyfus, G., Roussel, P., Stone, M., Denby, B. Robust contour tracking in the ultrasound tongue image sequences. Submitted to *Clinical linguistics & phonetics*.

[16] Zhou, X., Huang, X., Duncan, J. S., Yu, W. 2013. Active Contours with Group Similarity. *Proc. of IEEE CVPR* Oregon, 2969-2976.

[17] Choi, M. G., Ko, H. S. 2005. Modal warping: Real-time simulation of large rotational deformation and manipulation. *IEEE Transactions on Visualization and Computer Graphics*, 11(1), 91-101.

[18] Yang, Y., Guo, X., Vick, J., Torres, L. G., Campbell, T. F. 2013. Physics-based deformable tongue visualization. *IEEE Transactions on Visualization and Computer Graphics*, 19(5), 811-823.

[19] Jacob, M., Lehnert-LeHouillier, H., Bora, S., McAleavey, S., Dalecki, D., McDonough, J. Speckle tracking for the recovery of displacement and velocity information from sequences of ultrasound images of the tongue. *Proc. 8th ISSP* Strasburg, 53-56.

[20] Denby, B., Oussar, Y., Dreyfus, G., & Stone, M. (2006). Prospects for a Silent Speech interface using ultrasound imaging. *Proc. ICASSP* Toulouse, 365-368.

[21] Tang, L., Bressmann, T., Hamarneh, G. (2012). Tongue contour tracking in dynamic ultrasound via higher-order MRFs and efficient fusion moves. *Medical image analysis*, 16(8), 1503-1520.

[22] Xu, K., Yang, Y., Jaumard-Hakoun, A., Adda-Decker, M., Amelot, A., Crevier-Buchman, L., Chawah, P., Fux, T., Pillot-Loiseau, C., Al Kork, S. K., Roussel, P., Stone, M., Denby, B. (2014). 3D tongue motion visualization based on ultrasound image sequences. *Proc. 15th InterSpeech* Singapore, 1482-1483.